\documentclass{article}
\usepackage{conference,times}

\usepackage{amsmath,amsfonts,bm}

\def\eqref#1{equation~\ref{#1}}

\def\1{\bm{1}}

\DeclareMathAlphabet{\mathsfit}{\encodingdefault}{\sfdefault}{m}{sl}
\SetMathAlphabet{\mathsfit}{bold}{\encodingdefault}{\sfdefault}{bx}{n}

\usepackage{url}
\usepackage{graphicx}
\usepackage{wrapfig}
\usepackage{tabularx}
\usepackage{amssymb}
\usepackage{xcolor}
\usepackage{colortbl}
\usepackage{booktabs}
\usepackage{float}
\usepackage{microtype}
\usepackage{needspace}
\usepackage[hidelinks]{hyperref}
\usepackage[font=small,skip=6pt]{caption}
\graphicspath{{figures/}}
\definecolor{oursrow}{RGB}{232,245,233}
\iclrfinalcopy

\title{Foresight Without Seeing:\\
Latent Futures for World Action Models}

\author{
\textbf{Jiakai Huang$^{1,2}$} \quad
\textbf{Zhongbo Wu$^{1,2,3}$} \quad
\textbf{Siyu Xu$^{5}$} \quad
\textbf{Zheng Zhang$^{2,4}$} \quad
\textbf{Zihan Wang$^{1}$}\thanks{Research intern.}
\\
\ \textbf{Shan You$^{2}$} \quad
\ \textbf{Chang Xu$^{5}$} \quad
\textbf{Tao Huang$^{1}$}\thanks{Corresponding author.}
\\
$^{1}$Shanghai Jiao Tong University \quad
$^{2}$ACE Robotics \quad
$^{3}$Shanghai Innovation Institute 
\\
$^{4}$Nanyang Technological University \quad
$^{5}$University of Sydney 
}

\begin{document}
\raggedbottom

\maketitle

\vspace{-0.5in}
\begin{center}
\textcolor{blue}{\url{https://forewam.github.io/}}
\end{center}

\begin{abstract}
World Action Models (WAMs) connect visual prediction with robot control, yet supplying predictive context often comes at the cost of expensive future-video generation. Direct policies avoid this cost but lack an explicit interface for accessing future-indexed predictive information. Our key insight is that forecasting and rendering can be separated: a policy can access latent predictive computation without reconstructing future observations. Building on this insight, we introduce \textbf{ForeWAM}, a World Action Model that exposes and shapes latent predictive context for efficient control. Its core mechanism, \textbf{Future-KV}, performs a single Video DiT prefill over the current visual latent and noise-initialized future slots, then reuses the resulting key--value states throughout action denoising. Because predictive context alone does not ensure relevance to control, we further introduce dynamics registers supervised by latent actions from a frozen teacher during training, encouraging representations of interaction-induced transitions. Concentrating predictive computation in this reusable context supports a lightweight, single-layer action decoder. We evaluate ForeWAM on LIBERO, LIBERO-Plus, RoboCasa, and real-world manipulation tasks. Without additional policy-level embodied pretraining, ForeWAM improves RoboCasa success by 9.7 percentage points over Fast-WAM at the same budget of 50 demonstrations per task, reaching 59.2\%. With a single-layer decoder, it achieves 77.6\% success on LIBERO-Plus and reduces policy-query latency to 88.7\,ms on an NVIDIA A800, delivering a $6.27\times$ speedup over Fast-WAM. These results show that exposing and shaping latent predictive computation provides useful foresight for robust, efficient control without explicit future-video generation.
\end{abstract}

\section{Introduction}

\begin{wrapfigure}{r}{0.35\textwidth}
    \vspace{-16pt}
    \centering
    \setlength{\intextsep}{0pt}
    \includegraphics[width=\linewidth]{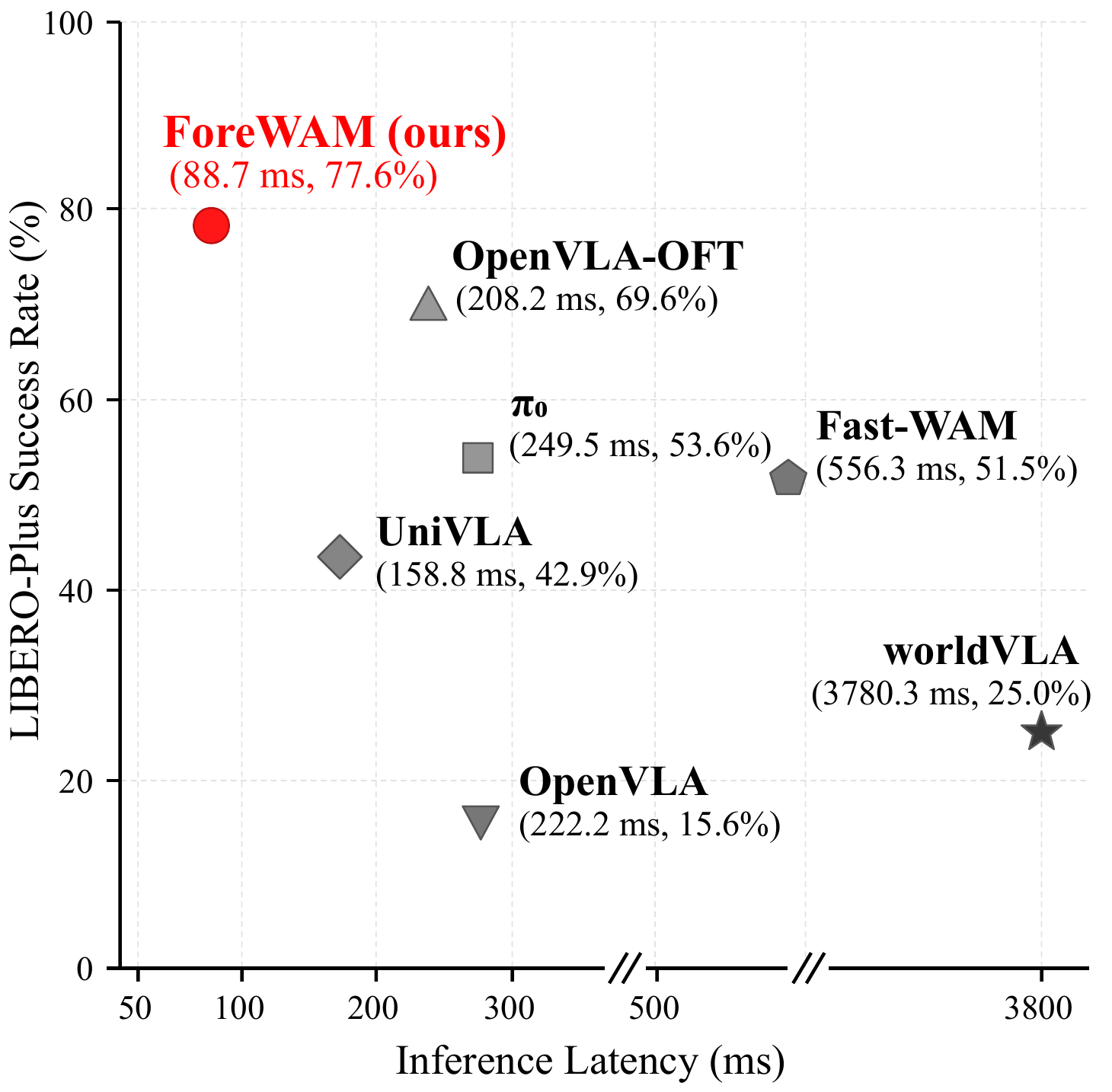}
    \setlength{\abovecaptionskip}{-3pt}
    \setlength{\belowcaptionskip}{-3pt}
    \caption{\textbf{OOD performance vs.\ latency on A800.}}
    \setlength{\intextsep}{-6pt}
    \label{fig:ood_latency}
    \vspace{-16pt}
\end{wrapfigure}

World Action Models (WAMs) couple visual world modeling with robot control so that action generation can benefit from predicted scene evolution~\citep{du2023learning,hu2024video,li2503unified,ye2026world,zhu2025unified,kairosteam2026kairosregretawarenativeworldaction}. Yet existing WAMs face a tension between predictive context and inference efficiency. Cascaded and joint WAMs explicitly expose predicted futures to the policy, but require costly iterative video generation~\citep{du2023learning,ye2026world,bi2026motus}. Direct-policy WAMs such as Fast-WAM~\citep{yuan2026fast} avoid this cost by acting directly on current-observation world-model features, but consequently provide no explicit interface for future-indexed predictive computation to reach the action pathway.

We argue that this tension stems from conflating \emph{forecasting} with \emph{rendering}. A future observation is only the final output of predictive computation: before a clean future frame is produced, a Video DiT already forms hidden states at future temporal positions from the current visual context, instruction, and robot state. For control, the policy need not recover the future in pixel space; it primarily needs the predictive cues encoded along the way. This suggests a different interface between world modeling and action generation: expose the latent computation of forecasting itself, rather than its rendered output. In other words, a policy may obtain \emph{foresight without seeing the future}.

\begin{figure}[t]
    \centering
    \includegraphics[width=\linewidth]{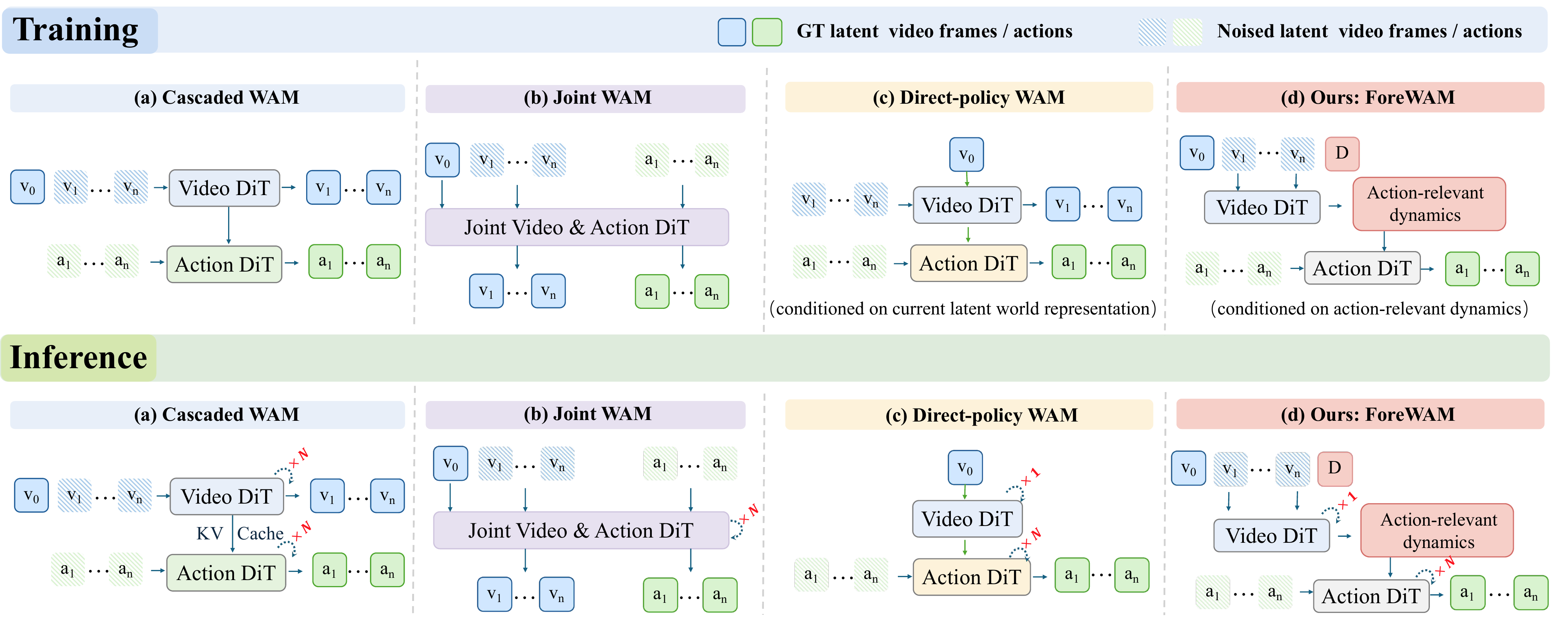}
    \caption{\textbf{World Action Model paradigms.}
    (a) Cascaded WAMs first generate future observations and then condition action prediction on them.
    (b) Joint WAMs generate future observations and actions within a unified generative process.
    (c) Direct-policy WAMs skip future rollout at inference and condition action prediction on a latent world representation extracted from the current observation.
    (d) ForeWAM retains direct action prediction while conditioning the action model on current-observation features, hidden future-slot K/V states, and dynamics registers trained with latent-action supervision.
    Hatched tokens denote noisy variables; future slots are stochastic internal states rather than observed future frames.}
    \label{fig:paradigm}
\end{figure}

Based on this insight, we introduce \textbf{ForeWAM}, a \textbf{Foresight-without-Seeing World Action Model}. Its core mechanism, \textbf{Future-KV}, exposes future-indexed predictive states without generating future video. ForeWAM appends noise-initialized future slots to the current visual latent and performs a single Video DiT prefill. The resulting key--value states from the current and future positions are cached and reused throughout action denoising. The action policy can therefore access predictive context computed by the video model, while avoiding iterative future-video denoising and decoding.

Yet predictive context is not necessarily \emph{action-relevant} context. Video prediction models broad scene evolution, including many changes that may be irrelevant to manipulation, whereas action generation depends particularly on interaction-induced transitions and task progress. We therefore introduce \textbf{dynamics registers} and supervise them with latent actions extracted by a frozen LaWM teacher~\citep{chen2026lawam} from demonstrated current--future observation pairs. This training-only signal shapes the exposed predictive representation toward dynamics that matter for control. The distinction is strongly reflected in our ablation: on LIBERO-Plus, Future-KV alone improves the matched baseline from 53.6\% to 58.5\%, and latent-action supervision alone reaches 54.8\%, whereas combining them yields 76.7\% (Table~\ref{tab:component_design}).

With predictive reasoning concentrated in the reusable video-side context, the action branch can be substantially simplified. The Video DiT is evaluated only once per policy query, whereas the Action DiT runs repeatedly throughout action denoising; shifting computation toward the former is therefore particularly beneficial for inference. We find that a single-layer Action DiT reaches 77.6\% on LIBERO-Plus, compared with 76.7\% for a 30-layer decoder, while a one-layer Fast-WAM reaches only 57.9\% (Table~\ref{tab:action_depth_design}). This suggests that the shallow decoder is enabled by the richer predictive context, rather than action decoding being intrinsically easy.

Experiments across LIBERO, LIBERO-Plus, RoboCasa, and real-world manipulation validate this design. Without additional policy-level embodied pretraining, ForeWAM achieves 59.2\% success on RoboCasa with only 50 demonstrations per task, outperforming Fast-WAM by 9.7 percentage points under the same demonstration budget. It retains near-saturated performance on standard LIBERO while substantially improving robustness under the distribution shifts of LIBERO-Plus. Meanwhile, ForeWAM reduces policy-query latency from 556.3\,ms to 88.7\,ms on an NVIDIA A800, a $6.27\times$ speedup over Fast-WAM (Figure~\ref{fig:ood_latency}). Together, these results show that useful world-model foresight need not be materialized as future video: it can instead be exposed and shaped directly in latent space for efficient control.

\section{Related Work}
\paragraph{Vision-language-action policies.}
VLA models map visual observations and language instructions to executable robot
actions~\citep{brohan2022rt,brohan2023rt,kim2024openvla,liu2025rdt,team2024octo,yang2026abot},
with many recent models adapting pretrained
vision-language backbones for control ~\citep{black2024pi_0,intelligence2025pi_,zhao2025cot}.
Diffusion and flow objectives support multimodal continuous
action generation~\citep{black2024pi_0,chi2025diffusion,lipman2022flow}, while large-scale robot
pretraining can improve transfer across tasks and embodiments
~\citep{bjorck2025gr00t,bu2025univla,zheng2026x}. However, the action-prediction objective alone does not explicitly require learning predictive representations of how the environment evolves during interaction, motivating the integration of world modeling into policy learning.

\paragraph{World-action models.}
World Action Models (WAMs) couple world prediction with action generation
through cascaded or joint architectures.
Cascaded approaches follow an \emph{imagine-then-act} pipeline, extracting actions
from generated future visual observations as intermediate
plans~\citep{du2023learning,du2024video},
geometric representations
~\citep{bharadhwaj2024track2act,ko2024learning,xu2024flow,
zhi20253dflowaction} or predictive latent features~\citep{hu2024video,huang2024ardup,lou2026mask,yan2026s}.
Joint WAMs learn future states and actions together. Autoregressive variants organize them within
a unified temporal sequence
~\citep{cen2025rynnvla,cen2025worldvla,cheang2024gr,wu2024unleashing,team2026kairos}.
Diffusion- and flow-based variants use either unified prediction backbones~\citep{chen2026unified,guo2024prediction,kim2026cosmos,shen2026videovla,ye2026world,
zhao2026frappe,zheng2025flare,zhu2025unified,world_model_benchmarks_survey_2026} or coordinated modality-specific streams, coupling their predictive streams through shared attention~\citep{bi2026motus,lyu2026lda,won2025dual,
yang2025covar,yuan2026adaworldpolicy}, intermediate hidden states~\citep{li2026world,ma2026dit4dit}, or shared representations
~\citep{
li2503unified}.
Iterative future generation and tightly coupled world--action computation
can incur substantial inference overhead. Direct-policy WAMs, including
Fast-WAM and GigaWorld-Policy, retain world-modeling supervision during
training but use action-only inference~\citep{ye2026gigaworld,yuan2026fast}.
Concurrent work explores complementary forms of predictive supervision:
Enfold~\citep{zeng2026enfold} distills world-generator states into
representations predicted from current observations and language, while
Robust-WAM~\citep{yan2026robust} aligns action-stream queries with future
visual semantics. ForeWAM retains noisy future slots at inference and
conditions action denoising on their hidden K/V states from a single Video
DiT prefill, together with dynamics registers supervised by LaWM
latent-action targets~\citep{chen2026lawam}, without generating future video.

\section{From Predictive Context to Actionable Foresight}
\label{sec:motivation}

To understand what an action policy actually needs from a video world model,
we conduct a controlled study on LIBERO-Plus using Fast-WAM retrained with
Wan2.1 as the matched direct-policy baseline~\citep{yuan2026fast}.
We focus on two questions: whether future-indexed latent context is sufficient
for control, and how the resulting context changes the capacity required by
the action pathway. The observations below motivate the design of ForeWAM in
Section~\ref{sec:method}.

\subsection{Useful Foresight Requires Both Access and Shaping}
\label{sec:motivation_context}

\begin{table}[!htbp]
    \centering
    \footnotesize
    \setlength{\tabcolsep}{3pt}
    \caption{\textbf{Predictive-context design study on LIBERO-Plus (\%).}
Future-KV exposes future-indexed context, while LA supervision shapes
interaction-relevant dynamics. All configurations use a 30-layer Action DiT
under matched settings. Bold marks the best result in each column.}
    \label{tab:component_design}
    \begin{tabular*}{\linewidth}{@{\extracolsep{\fill}}cc|cccccccc@{}}
        \toprule
        LA & Future-KV & Camera & Robot & Language & Light
        & Background & Noise & Layout & Avg \\
        \midrule
        $\times$ & $\times$
        & 37.0 & 36.5 & 64.1 & 81.5 & 49.3 & 50.1 & 63.7 & 53.6 \\
        $\checkmark$ & $\times$
        & 49.0 & 36.8 & 65.1 & 75.9 & 48.5 & 50.8 & 61.7 & 54.8 \\
        $\times$ & $\checkmark$
        & 37.8 & 41.0 & 63.9 & 83.4 & 65.7 & 61.5 & 65.6 & 58.5 \\
        $\checkmark$ & $\checkmark$
        & \textbf{64.6} & \textbf{61.6} & \textbf{86.6} & \textbf{95.3}
        & \textbf{74.1} & \textbf{77.7} & \textbf{81.7} & \textbf{76.7} \\
        \bottomrule
    \end{tabular*}
\end{table}

We first examine how predictive information should be exposed to the action
policy. The matched Fast-WAM baseline conditions action generation on
current-observation world-model features and achieves 53.6\% average success.
Providing additional future-indexed context through Future-KV improves this
result to 58.5\% (Table~\ref{tab:component_design}), despite requiring no
explicitly generated future observation at inference. This suggests that
internal states associated with future temporal positions provide useful
context beyond current-observation features alone.

However, access to such context is clearly not sufficient. Video prediction
models scene evolution broadly, whereas control depends particularly on
interaction-induced transitions and task progress. Consistent with this
distinction, latent-action (LA) supervision alone provides only a marginal
gain, from 53.6\% to 54.8\%, while Future-KV alone reaches 58.5\%.
Strikingly, combining the two increases success to 76.7\%, outperforming the
stronger individual component by 18.2 percentage points. The same
combination improves all seven LIBERO-Plus perturbation categories over
either component alone
(Table~\ref{tab:component_details} in
Appendix~\ref{app:component_details}).

These results reveal two complementary requirements for useful foresight.
Future-KV determines \emph{what predictive context is accessible} to the
action pathway, whereas LA supervision shapes \emph{what information that
context emphasizes}. In other words, predictive context becomes substantially
more useful for control when future-indexed representations are both
\textbf{exposed} and \textbf{shaped toward interaction-relevant dynamics}.

\subsection{Rich Predictive Context Enables Lightweight Action Decoding}
\label{sec:motivation_depth}

The representation supplied by the video model also determines how much
computation remains necessary in the action pathway. The Video DiT constructs
context once per policy query, whereas the Action DiT is evaluated repeatedly
throughout action denoising. If the video-side context already captures the
dynamics needed for control, a deep action decoder may therefore be redundant.

\begin{table}[!htbp]
    \centering
    \footnotesize
    \setlength{\tabcolsep}{3pt}
    \caption{Action-expert depth comparison. LIBERO-Plus success rate (\%),
    without additional policy-level embodied pretraining. All configurations retain the 30-layer
    Wan2.1 Video DiT; action-layer mappings and initializations follow
    Appendix~\ref{app:training_details}. Bold marks the best result in each
    column.}
    \label{tab:action_depth_design}
    \begin{tabular}{@{}ll|cccccccc@{}}
        \toprule
        \multicolumn{2}{c|}{Action DiT depth} & Camera & Robot & Language & Light
        & Background & Noise & Layout & Avg \\
        \midrule
        \multicolumn{2}{l|}{Fast-WAM / Wan2.1 (30 layers)}
        & 37.0 & 36.5 & 64.1 & 81.5 & 49.3 & 50.1 & 63.7 & 53.6 \\
        \multicolumn{2}{l|}{Fast-WAM / Wan2.1 (1 layer)}
        & 44.9 & 38.0 & 71.5 & 81.6 & 51.3 & 54.2 & 69.0 & 57.9 \\
        \midrule
        \multicolumn{2}{l|}{ForeWAM (30 layers)}
        & 64.6 & 61.6 & \textbf{86.6} & 95.3
        & 74.1 & 77.7 & 81.7 & 76.7 \\
        \multicolumn{2}{l|}{ForeWAM (12 layers)}
        & \textbf{67.7} & \textbf{63.7} & 86.0 & 90.8
        & 71.8 & 80.0 & 80.5 & 76.9 \\
        \multicolumn{2}{l|}{ForeWAM (6 layers)}
        & 61.5 & 61.4 & 79.6 & 92.2
        & 76.4 & 78.6 & 78.4 & 74.7 \\
        \multicolumn{2}{l|}{ForeWAM (1 layer)}
        & 61.7 & 56.8 & 85.1 & \textbf{95.8}
        & \textbf{76.5} & \textbf{89.1} & \textbf{83.0} & \textbf{77.6} \\
        \bottomrule
    \end{tabular}
\end{table}

We test this hypothesis by retaining all 30 layers of the Video DiT while
varying the Action DiT depth. ForeWAM achieves 76.7\%, 76.9\%, 74.7\%, and
77.6\% average success with 30, 12, 6, and 1 action layers, respectively
(Table~\ref{tab:action_depth_design}). Increasing action depth provides no
monotonic benefit; remarkably, the single-layer decoder achieves the highest
overall success among these variants.

This result is not simply explained by action decoding being intrinsically
easy. Reducing the matched Fast-WAM action branch from 30 layers to one
improves its success only from 53.6\% to 57.9\%. At the same one-layer depth,
ForeWAM reaches 77.6\%, a 19.7-point advantage. The lightweight decoder is
therefore effective in conjunction with the richer predictive context, rather
than because the original action pathway is unnecessarily deep.

Together, these observations suggest a clear division of computation:
the video branch should expose and organize predictive dynamics into an
action-relevant representation, while the repeatedly evaluated action branch
can remain lightweight. ForeWAM instantiates this principle with Future-KV,
latent-action-supervised dynamics registers, and a shallow Action DiT, which
we detail next.

\section{ForeWAM}
\label{sec:method}

\begin{figure}[t]
    \centering
    \includegraphics[width=\linewidth]{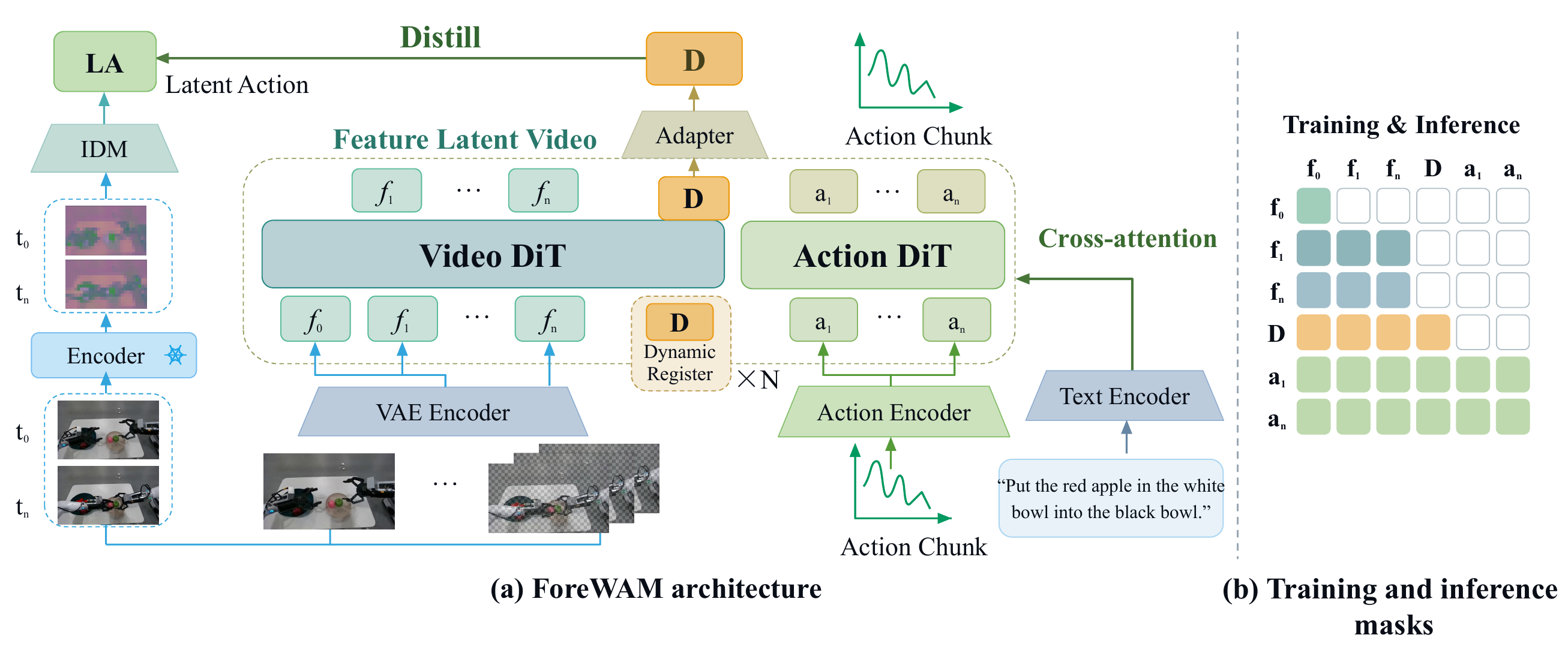}
    \setlength{\abovecaptionskip}{-8pt}
    \caption{\textbf{ForeWAM.}
    (a) Architecture and training pipeline. A supervision pass uses
    demonstrated futures for video and latent-action learning, while an
    independent action-conditioning pass uses pure-noise future slots to
    construct the predictive context consumed by the action policy.
    At inference, only the latter is required: one Video DiT prefill supplies
    cached K/V throughout action denoising, without future observations or
    the latent-action teacher.
    (b) Structured attention mask shared by training and inference.
    Rows are queries and columns are keys; colored cells indicate allowed
    attention. Here $f_0=C$, $f_{1:n}=F$, and $a_{1:n}=A$.
    Future slots attend bidirectionally to one another.}
    \label{fig:architecture}
\end{figure}

Guided by the observations in Section~\ref{sec:motivation}, we introduce
\textbf{ForeWAM}, a direct-policy World Action Model that exposes predictive
context from a video world model without explicitly generating future video.
ForeWAM follows a simple decomposition. \textbf{Future-KV} extracts
future-indexed latent context through a single Video DiT prefill; learnable
\textbf{dynamics registers} shape this context toward interaction-relevant
transitions through latent-action supervision; and a lightweight Action DiT
reuses the resulting predictive context throughout action denoising.
Figure~\ref{fig:architecture} illustrates the overall architecture.

\paragraph{Problem formulation.}
\label{sec:problem_formulation}
Given synchronized multi-camera observations $o$, a language instruction $l$,
and proprioceptive state $p$, the policy predicts an executable action chunk
$a_{1:H}\in\mathbb{R}^{H\times d_{\mathrm{act}}}$.
We retain the direct-policy interface
$p_\theta(a_{1:H}\mid o,l,p)$~\citep{yuan2026fast}.
Let $z_{\mathrm{cur}}$ denote the current-frame latent encoded by the video
VAE, and $z_{1:T}$ the demonstrated video latents used for training
supervision. Future observations and latent-action teacher outputs are used
only during training.

\subsection{Latent Foresight through Future-KV}
\label{sec:model_architecture}
\label{sec:future_kv}

The first component exposes predictive computation associated with future
temporal positions without materializing future observations. For the
action-conditioning pathway, we concatenate the clean current latent with
Gaussian future slots:
\begin{equation}
    z^{\mathrm{init}}
    = [z_{\mathrm{cur}};\epsilon_F],
    \qquad \epsilon_F\sim\mathcal{N}(0,I),
    \label{eq:future_substrate}
\end{equation}
where $[\cdot;\cdot]$ denotes temporal concatenation.
The Video DiT processes this sequence once at $\sigma=1.0$, conditioned on
the instruction and proprioception. We retain the resulting hidden
key--value states and reuse them throughout action denoising. We refer to
this interface as \textbf{Future-KV}.

Importantly, the future slots contain no observed future information.
Instead, they provide future temporal positions at which a video model
trained with future supervision can construct contextual states from the
available inputs. ForeWAM exposes these internal states directly to the
action pathway rather than iteratively denoising them into valid future
frames. Thus, predictive context is computed once and reused, while explicit
future-video generation is avoided.

\subsection{Shaping Predictive Context with Latent Actions}
\label{sec:latent_action_design}

Future-KV provides access to future-indexed context, but does not by itself
specify which aspects of scene evolution should be emphasized for control.
We therefore introduce learnable dynamics registers
$D=\{D_i\}_{i=1}^{N_D}$ into the Video DiT and supervise their pooled
representation with a latent-action target extracted from demonstrated
transitions.

\paragraph{Latent-action supervision.}
We use a frozen LaWAM teacher~\citep{chen2026lawam} to encode a demonstrated
transition. Let $o_{\mathrm{ext}}$ and $o_{\mathrm{ext}}^+$ denote the current
and future observations from the external-camera view. A frozen visual
encoder $f_{\mathrm{vis}}$ and latent-action extraction mapping
$E_{\mathrm{IDM}}$ produce
\begin{equation}
    \begin{gathered}
    u_{\mathrm{cur}}=f_{\mathrm{vis}}(o_{\mathrm{ext}}),\quad
    u_{\mathrm{fut}}=f_{\mathrm{vis}}(o_{\mathrm{ext}}^+),\\
    z_{\mathrm{LA}}
    =E_{\mathrm{IDM}}(u_{\mathrm{cur}},u_{\mathrm{fut}}).
    \end{gathered}
    \label{eq:latent_action_target}
\end{equation}
The teacher representation $z_{\mathrm{LA}}$ describes the demonstrated
transition and is used only as a training target. It is distinct from the
video VAE latents $z_{1:T}$ and is never supplied directly to the action
decoder. A trainable projection $g_\psi$ aligns the pooled dynamics-register
representation with this latent-action target.

\paragraph{Routing predictive context.}
ForeWAM organizes information flow among current-frame tokens $C$,
future-slot tokens $F$, dynamics registers $D$, and action tokens $A$ using
the structured attention mask in Figure~\ref{fig:architecture}(b).
Within the video branch, current-frame tokens attend only to $C$, while
future-slot tokens attend to $C$ and all tokens in $F$, with bidirectional
attention across future slots. Dynamics registers attend to $C$, $F$, and
$D$, allowing them to aggregate current and future-indexed information
without feeding back into the visual tokens.

The resulting video-side context is exposed to the action branch through
cached K/V. At each action-denoising step, action queries attend to cached
states from $C$, $F$, and $D$, together with the current action-token K/V
from $A$. Both branches additionally receive text embeddings and projected
proprioceptive conditioning, which are omitted from the mask diagram for
clarity.

\subsection{Lightweight Action Decoding from Cached Predictive Context}
\label{sec:action_depth}

As shown in Section~\ref{sec:motivation_depth}, once the video branch
provides rich predictive context, the action pathway can be substantially
simplified. We therefore use a single-layer Action DiT that reads the
final Video DiT layer, $L=30$.

A single video prefill produces the dynamics-register states and final-layer
K/V cache:
\begin{equation}
    (D,K,V)
    = \operatorname{Prefill}_\phi(z^{\mathrm{init}},l,p),
    \label{eq:future_kv_context}
\end{equation}
where $K=K^{(L)}$ and $V=V^{(L)}$ contain the final-layer states of the
current-frame, future-slot, and dynamics-register tokens. These features
condition action prediction as
\begin{equation}
    p_\theta(a_{1:H}\mid o,l,p,D,K,V).
    \label{eq:ours_policy}
\end{equation}
The dependence of $(D,K,V)$ on $(o,l,p,\epsilon_F)$ is implicit.

At inference, the Video DiT is evaluated only once for each action-chunk prediction.
During subsequent action denoising, the single-layer Action DiT repeatedly
reads the same cached predictive context together with the evolving action
tokens. ForeWAM therefore concentrates visual and predictive computation in
a reusable video-side representation while keeping repeated action-side
computation lightweight.

\subsection{Training and Inference}
\label{sec:training_objective}

Training uses two Video DiT forward passes with shared parameters but
independent hidden states and K/V caches. The \emph{supervision pass} receives
noised demonstrated futures and provides the video-prediction and
latent-action objectives. The independent \emph{action-conditioning pass}
instead receives freshly sampled pure-noise future slots at $\sigma=1.0$,
matching the inference-time input in Eq.~\ref{eq:future_substrate}, and
produces $(D,K,V)$ for the action objective. This separation ensures that
demonstrated future observations supervise the model during training without
being directly exposed to the action-conditioning pathway.

\paragraph{Video flow matching.}
The video branch learns demonstrated visual dynamics through continuous flow
matching~\citep{lipman2022flow}. Writing the target video latents as
$z=z_{1:T}$, we sample Gaussian noise $\epsilon_z$ and flow time
$\tau\in(0,1)$ to construct
$z_\tau=(1-\tau)z+\tau\epsilon_z$, while keeping the current-frame latent
clean. The video velocity predictor $v_\phi$ is optimized with
\begin{equation}
    \mathcal{L}_{\mathrm{video}}
    =\mathbb{E}_{z,\epsilon_z,\tau}
    \left[
        \left\|
        v_\phi(z_\tau,\tau\mid o,l,p)-(\epsilon_z-z)
        \right\|_2^2
    \right].
    \label{eq:flow_matching}
\end{equation}
Let $D^{\mathrm{sup}}$ denote the dynamics-register states produced by this
supervision pass.

\paragraph{Action flow matching.}
Let $a=a_{1:H}$ be the demonstrated action chunk. We sample Gaussian noise
$\epsilon_a$ and action flow time $s\in(0,1)$, and form
$a_s=(1-s)a+s\epsilon_a$. Conditioned on the predictive context
$(D,K,V)$ from the independent pure-noise prefill, the Action DiT predicts
the action velocity:
\begin{equation}
    \mathcal{L}_{\mathrm{action}}
    =\mathbb{E}
    \left[
        \left\|
        v_\theta(a_s,s\mid o,l,p,D,K,V)-(\epsilon_a-a)
        \right\|_2^2
    \right].
    \label{eq:action_flow_matching}
\end{equation}
The expectation is over training examples, action noise and flow time, and
fresh future-slot noise $\epsilon_F$. Crucially, the noised demonstrated
future $z_\tau$, the supervision-pass registers $D^{\mathrm{sup}}$, and their
K/V states never enter this action-conditioning forward pass.

\paragraph{Latent-action supervision.}
The projection $g_\psi$ aligns the mean-pooled supervision-pass dynamics
registers with the frozen latent-action target from
Eq.~\ref{eq:latent_action_target}:
\begin{equation}
    \mathcal{L}_{\mathrm{LA}}
    =\left\|
    g_\psi\!\left(
        \frac{1}{N_D}\sum_{i=1}^{N_D}D_i^{\mathrm{sup}}
    \right)
    -\operatorname{sg}(z_{\mathrm{LA}})
    \right\|_2^2,
    \label{eq:latent_action_distillation}
\end{equation}
where $\operatorname{sg}$ stops gradients through the frozen teacher.
The complete training objective is
\begin{equation}
    \mathcal{L}
    =\mathcal{L}_{\mathrm{video}}
    +\mathcal{L}_{\mathrm{action}}
    +\lambda_{\mathrm{LA}}\mathcal{L}_{\mathrm{LA}}.
    \label{eq:total_objective}
\end{equation}
Action gradients propagate through the video-to-action K/V interface,
allowing the predictive context to be optimized jointly for control.

\paragraph{Inference.}
At inference, the supervision pass, demonstrated future observations, and
latent-action teacher are all removed. ForeWAM encodes the current
observation, samples pure-noise future slots, performs a single Video DiT
prefill, and reuses the resulting $(D,K,V)$ cache throughout action
denoising. No future video is generated or decoded.

\section{Experiments}
\subsection{Experimental Setup}
\label{sec:experimental_setup}

We evaluate ForeWAM on LIBERO, RoboCasa, the 
OOD LIBERO-Plus benchmark, and real-world 
manipulation tasks, using success rate as the primary metric.

\paragraph{Benchmarks and protocols.}
We train on the original LIBERO demonstrations (500 per suite) and evaluate
on its four standard suites and the seven LIBERO-Plus perturbation categories,
without using augmented LIBERO-Plus training data~\citep{liu2023libero,fei2025libero}.
We evaluate RoboCasa on 24 static
kitchen tasks over three seeds (3,600 trials), using successful human
demonstrations without additional MimicGen data.

\paragraph{Real-world experiments.}
We jointly train on 125 teleoperated demonstrations (71,702 frames) from four
tabletop tasks using a 6-DoF AgileX Piper arm with third-person and wrist RGB
cameras. ID evaluation uses 10 trials per task with matched randomized object
placements; OOD evaluation perturbs object attributes in pen placement and
drawer manipulation (Figure~\ref{fig:real_robot}). Full benchmark protocols
and task definitions appear in Appendix~\ref{app:evaluation_protocols};
additional real-world details are in Appendix~\ref{app:real_world_details}.

\subsection{Main Results}

\begin{table}[!htbp]
    \centering
    \begin{minipage}[t]{0.51\linewidth}
        \vspace{0pt}
        \centering
        \caption{Results on RoboCasa.}
        \label{tab:robocasa_results}
    \footnotesize
    \setlength{\tabcolsep}{1.5pt}
    \renewcommand{\arraystretch}{1.08}
    \begin{tabularx}{\linewidth}{@{}>{\raggedright\arraybackslash}X|c|c|c@{}}
        \toprule
        Method & P.T. & \shortstack{Demos/\\task} & \textbf{SR (\%)} \\
        \midrule
        GR00T-N1~\citep{bjorck2025gr00t} & $\checkmark$ & 300 & 49.6 \\
        DP-VLA~\citep{han2024dual} & $\checkmark$ & 3000 & 57.3 \\
        GR00T-N1 + DreamGen~\citep{jang2025dreamgen} & $\checkmark$ & 300 & 57.6 \\
        GR00T-N1 + DUST~\citep{won2025dual} & $\checkmark$ & 300 & 58.5 \\
        
        UVA~\citep{li2503unified} & $\times$ & 50 & 50.0 \\
        Fast-WAM~\citep{yuan2026fast} & $\times$ & 50 & 49.5 \\
        \rowcolor{oursrow}
        \textbf{ForeWAM} & $\times$ & 50 & \textbf{59.2} \\
        \bottomrule
    \end{tabularx}
    \end{minipage}\hfill
    \begin{minipage}[t]{0.47\linewidth}
        \vspace{0pt}
        \centering
        \caption{Results on LIBERO.}
        \label{tab:libero_results}
    \footnotesize
    \setlength{\tabcolsep}{1.5pt}
    \renewcommand{\arraystretch}{1.08}
    \begin{tabularx}{\linewidth}{@{}>{\raggedright\arraybackslash}X|c|c@{}}
        \toprule
        Method & P.T. & \textbf{Avg.} \\
        \midrule
        OpenVLA~\citep{kim2024openvla} & $\checkmark$ & 76.5 \\
        WorldVLA~\citep{cen2025worldvla} & $\checkmark$ & 81.8 \\
        $\pi_0$-Fast~\citep{pertsch2025fast} & $\checkmark$ & 85.5 \\
        $\pi_0$~\citep{black2024pi_0} & $\checkmark$ & 94.1 \\
        UniVLA~\citep{bu2025univla} & $\checkmark$ & 95.2 \\
        $\pi_{0.5}$~\citep{intelligence2025pi_} & $\checkmark$ & 96.9 \\
        CogVLA~\citep{li2026cogvla} & $\checkmark$ & 97.4 \\
        Fast-WAM~\citep{yuan2026fast} & $\times$ & \underline{97.6} \\
        \rowcolor{oursrow}
        \textbf{ForeWAM} & $\times$ & \textbf{97.9} \\
        \bottomrule
    \end{tabularx}
    \end{minipage}
    \par\smallskip
    {\footnotesize\raggedright P.T.: policy-level embodied pretraining (Appendix~\ref{app:training_details}).\par}
\end{table}

\paragraph{Results on RoboCasa.}
ForeWAM achieves strong manipulation performance with a limited demonstration
budget (Table~\ref{tab:robocasa_results}). With 50 demonstrations per task and
no additional policy-level embodied pretraining, it reaches 59.2\% average success across 24 kitchen
manipulation tasks, compared with 49.5\% for Fast-WAM under the same
demonstration budget and pretraining setting, a gain of 9.7 percentage points.
ForeWAM also exceeds GR00T-N1~\citep{bjorck2025gr00t} by 9.6 percentage points while using one-sixth
as many demonstrations per task (50 versus 300), achieving the highest
success rate among the methods listed in the table.

\paragraph{Results on LIBERO \& LIBERO-Plus.}
ForeWAM preserves the near-saturated performance of Fast-WAM on standard
LIBERO while substantially improving robustness under the distribution shifts
evaluated by LIBERO-Plus
(Tables~\ref{tab:libero_results} and~\ref{tab:libero_plus_results}).
On LIBERO, ForeWAM achieves 97.9\% average success compared with 97.6\%
for Fast-WAM, with neither method using additional policy-level embodied pretraining.
Suite-level results are provided in Appendix~\ref{app:libero_results}.
On the full LIBERO-Plus benchmark, the improvement is much larger:
ForeWAM raises average success from 51.5\% to 77.6\%, a gain of 26.1
percentage points, and exceeds OpenVLA-OFT by 8.0 points.
It leads five of the seven perturbation categories, including lighting
(95.8\%) and sensor noise (89.1\%). These results show that the main gain
over Fast-WAM lies in robustness to the evaluated distribution shifts,
while standard manipulation performance remains high.

\begin{table}[!htbp]
    \centering
    \footnotesize
    \setlength{\tabcolsep}{1.5pt}
    \caption{Success rates (\%) on seven LIBERO-Plus
    perturbation categories.}
    \label{tab:libero_plus_results}

    \begin{tabularx}{\linewidth}{@{}>{\raggedright\arraybackslash}Xcccccccccc@{}}
        \toprule
        Method & Params & P.T. & Camera & Robot & Language
        & Light & Background & Noise & Layout & Avg \\
        \midrule

        OpenVLA~\citep{kim2024openvla}
        & 7B & $\checkmark$
        & 0.8 & 3.5 & 23.0 & 8.1 & 34.8 & 15.2 & 28.5 & 15.6 \\

        OpenVLA-OFT
        & 7.5B & $\checkmark$
        & 56.4 & 31.9 & \underline{79.5} & \underline{88.7} & \textbf{93.3}
        & 75.8 & \underline{74.2} & \underline{69.6} \\

        $\pi_0$~\citep{black2024pi_0}
        & 3.3B & $\checkmark$
        & 13.8 & 6.0 & 58.8 & 85.0 & \underline{81.4}
        & \underline{79.0} & 68.9 & 53.6 \\


        $\pi_0$-Fast~\citep{pertsch2025fast}
        & 3.3B & $\checkmark$
        & \textbf{65.1} & 21.6 & 61.0 & 73.2 & 73.2
        & 74.4 & 68.8 & 61.6 \\

        UniVLA~\citep{bu2025univla}
        & 7B & $\checkmark$
        & 1.8 & \underline{46.2} & 69.6 & 69.0
        & 81.0 & 21.2 & 31.9 & 42.9 \\

        WorldVLA~\citep{cen2025worldvla}
        & 7B & $\checkmark$
        & 0.1 & 27.9 & 41.6 & 43.7 & 17.1 & 10.9 & 38.0 & 25.0 \\

        Fast-WAM~\citep{yuan2026fast}
        & 6B & $\times$
        & 16.4 & 44.5 & 68.9 & 78.2 & 53.7
        & 37.7 & 60.7 & 51.5 \\

        \midrule

        \rowcolor{oursrow}
        ForeWAM
        & 1.5B & $\times$
        & \underline{61.7} & \textbf{56.8} & \textbf{85.1}
        & \textbf{95.8} & 76.5 & \textbf{89.1}
        & \textbf{83.0} & \textbf{77.6} \\


        \bottomrule
    \end{tabularx}
\end{table}

\begin{figure}[t]
    \centering
    \includegraphics[width=0.90\linewidth]{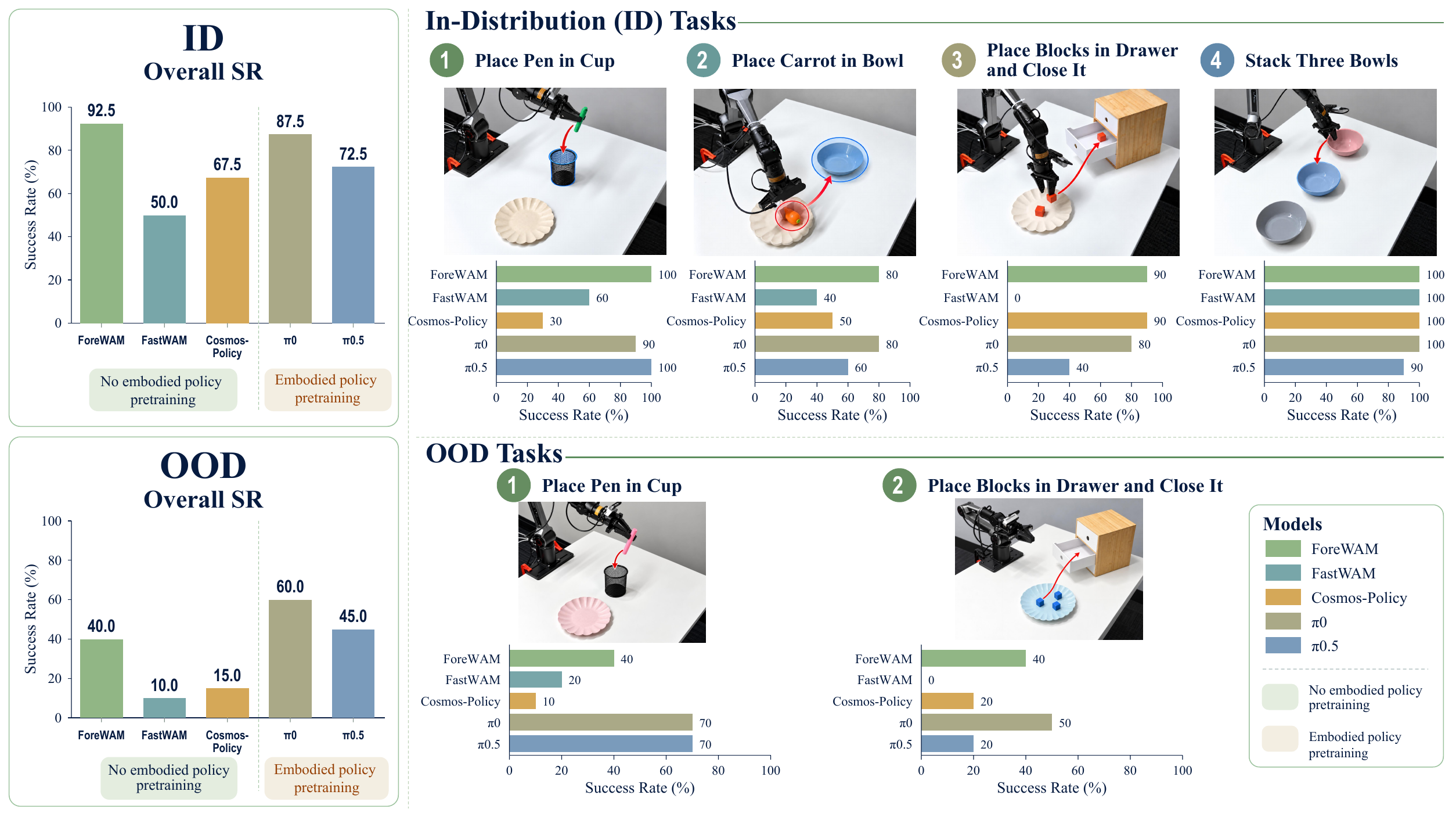}
    \caption{Real-world task success rates (\%). Top: four in-distribution
    tasks. Bottom: OOD evaluation with perturbed object attributes in pen
    placement and drawer manipulation.
    Left panels report the mean success rate across the tasks in each setting;
    right panels show task examples and per-task results. ForeWAM and
    Fast-WAM use no additional policy-level embodied pretraining.}
    \label{fig:real_robot}
\end{figure}

\paragraph{Results on Real-World Tasks.}
ForeWAM achieves the highest average success rate across the four
in-distribution (ID) real-world tasks (Figure~\ref{fig:real_robot}).
Without additional policy-level embodied pretraining, it reaches 92.5\% success, outperforming
Fast-WAM by 42.5 percentage points and $\pi_0$ by 5.0 points.
The largest gain over Fast-WAM occurs on placing blocks in the drawer
and closing it, where success increases from 0\% to 90\%.
Under object-attribute perturbations in the two OOD tasks, ForeWAM achieves 40.0\%
average success, compared with 10.0\% for Fast-WAM and 15.0\% for
Cosmos Policy~\citep{kim2026cosmos}. While ForeWAM outperforms both $\pi_0$ and $\pi_{0.5}$ on ID tasks,
all three evaluated WAMs remain below $\pi_0$ (60.0\%) and
$\pi_{0.5}$ (45.0\%) in average OOD success. This shared gap may
partly reflect differences in pretraining, as ForeWAM, Fast-WAM,
and Cosmos Policy do not undergo the large-scale policy-level
robot pretraining used by these two policies.

\subsection{Analysis}

\begin{figure}[t]
    \centering
    \includegraphics[width=0.85\linewidth]{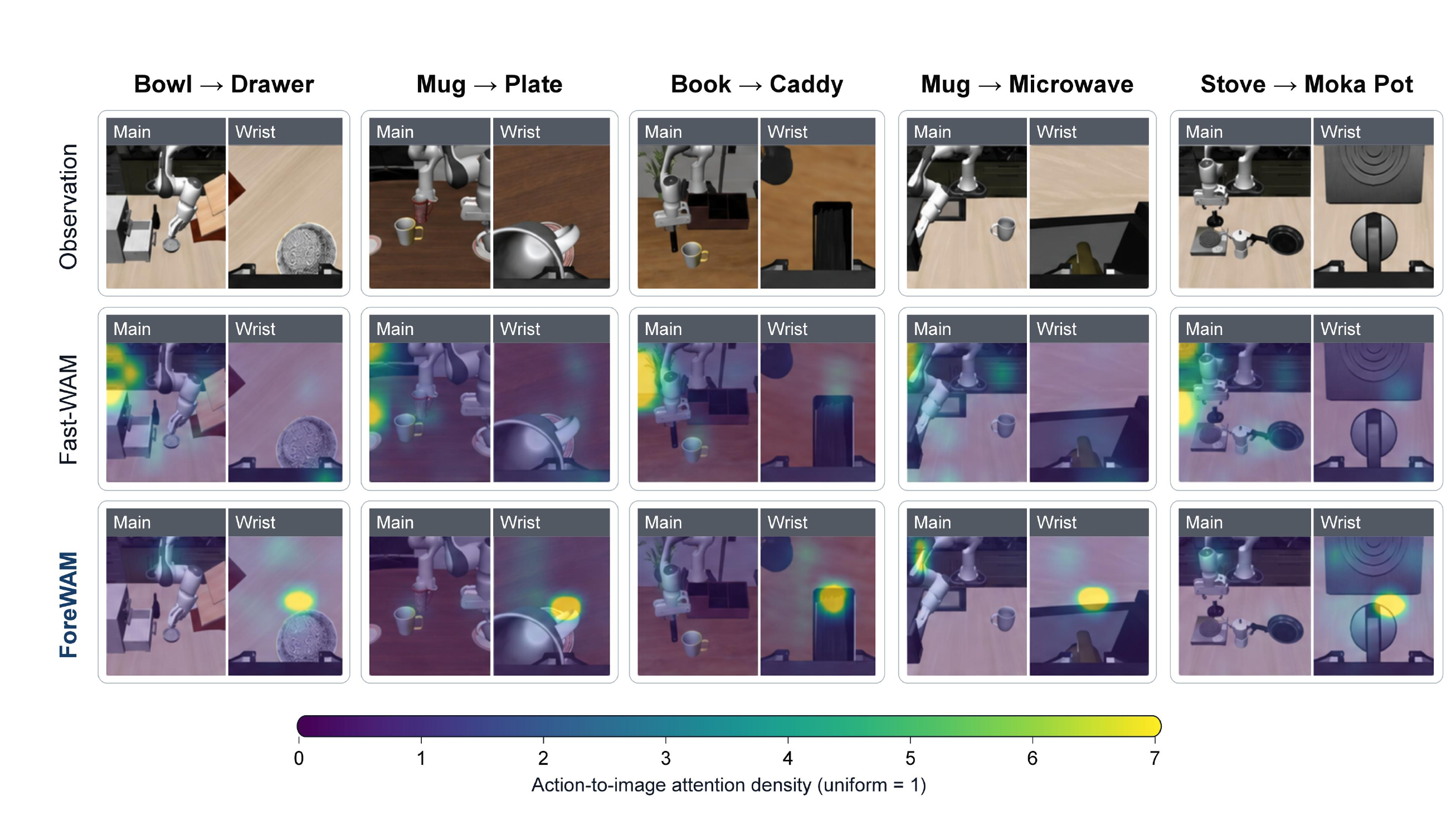}
    \caption{Attention maps.}
    \label{fig:attention_map}
\end{figure}

\paragraph{Action-to-image Attention.}
Figure~\ref{fig:attention_map} visualizes the
action-to-image attention maps of Fast-WAM and ForeWAM
across two manipulation examples. ForeWAM exhibits
more localized attention around interaction-relevant
regions, including manipulated objects, target receptacles,
and contact areas, particularly in wrist views, while
Fast-WAM shows more diffuse responses and stronger
activations in less task-relevant regions. This indicates
that ForeWAM provides more interaction-focused visual
context to the action pathway.

\Needspace{16\baselineskip}
\begin{wraptable}{r}{0.43\textwidth}
    \centering
    \footnotesize
    \setlength{\tabcolsep}{3pt}
    \caption{Inference latency (ms) on LIBERO-10.}
    \label{tab:inference_latency}
    \begin{tabular*}{\linewidth}{@{\extracolsep{\fill}}lrr@{}}
        \toprule
        Stage & Fast-WAM & ForeWAM \\
        \midrule
        VAE Enc. & 12.7 & \textbf{10.0} \\
        Visual & 47.7 & \textbf{46.4} \\
        Action & 470.6 & \textbf{21.7} \\
        Other & 25.3 & \textbf{10.6} \\
        \midrule
        \rowcolor{oursrow}
        Overall & 556.3 & \textbf{88.7} \\
        \bottomrule
    \end{tabular*}
\end{wraptable}

\paragraph{Inference Efficiency.}
Table~\ref{tab:inference_latency} compares overall and component-wise
inference latency on LIBERO-10 using a single NVIDIA A800 GPU and 10
action-denoising steps. ForeWAM reduces overall latency from 556.3\,ms to
88.7\,ms, yielding a $6.27\times$ speedup over Fast-WAM.

The largest reduction occurs in action generation, which decreases from
470.6\,ms to 21.7\,ms, while visual processing remains comparable
(46.4\,ms versus 47.7\,ms). Thus, the measured speedup primarily comes
from the action-generation stage. ForeWAM's single Video DiT prefill supplies
cached Future-KV features to the one-layer Action DiT throughout denoising,
retaining latent future conditioning with a lightweight action pathway.
Together with the success-rate results above, these measurements show that
ForeWAM combines strong manipulation performance with low inference latency.
Appendix~\ref{app:latency_comparison} provides details of the latency comparison.



\section{Limitations and Discussion}

Our evaluation covers LIBERO, LIBERO-Plus, RoboCasa, and real-world tabletop manipulation. However, the real-world experiments are limited to a single robot platform, four tasks, and a small number of trials. Generalization under the evaluated object-attribute shifts remains challenging, while transfer to other embodiments and broader task distributions has yet to be established. Extending the evaluation to more diverse robot platforms, interaction dynamics, and longer-horizon tasks is an important direction for future work.

\label{sec:limitations}

\section{Conclusion}

We introduced ForeWAM, a direct-policy World Action Model that provides predictive context for control without explicit future-video generation. ForeWAM combines Future-KV with latent-action-supervised dynamics registers: a single Video DiT prefill supplies cached context that a lightweight Action DiT reuses throughout action denoising. Future observations and the latent-action teacher are required only during training.

Experiments across simulation benchmarks and real-world manipulation tasks demonstrate the effectiveness of this design, while component ablations show the complementary benefits of Future-KV and latent-action supervision. With a single-layer action decoder, ForeWAM achieves 77.6\% success on LIBERO-Plus and a policy-query latency of 88.7\,ms on an NVIDIA A800. These results support exposing and shaping latent predictive context as an effective approach to robust, efficient control without rendering future observations.



\bibliographystyle{conference}
\begingroup
\small
\setlength{\bibsep}{3pt plus 0.3pt}
\bibliography{conference}

@article{du2023learning,
  title={Learning universal policies via text-guided video generation},
  author={Du, Yilun and Yang, Sherry and Dai, Bo and Dai, Hanjun and Nachum, Ofir and Tenenbaum, Josh and Schuurmans, Dale and Abbeel, Pieter},
  journal={Advances in neural information processing systems},
  volume={36},
  pages={9156--9172},
  year={2023}
}

@article{hu2024video,
  title={Video prediction policy: A generalist robot policy with predictive visual representations},
  author={Hu, Yucheng and Guo, Yanjiang and Wang, Pengchao and Chen, Xiaoyu and Wang, Yen-Jen and Zhang, Jianke and Sreenath, Koushil and Lu, Chaochao and Chen, Jianyu},
  journal={arXiv preprint arXiv:2412.14803},
  year={2024}
}

@article{wan2025wan,
  title={Wan: Open and advanced large-scale video generative models},
  author={Wan, Team and Wang, Ang and Ai, Baole and Wen, Bin and Mao, Chaojie and Xie, Chen-Wei and Chen, Di and Yu, Feiwu and Zhao, Haiming and Yang, Jianxiao and others},
  journal={arXiv preprint arXiv:2503.20314},
  year={2025}
}

@article{zeng2026enfold,
  title={Enfold: Folding World Model Imagination into Predictive Representations for Ultra-Efficient Embodied Control},
  author={Zeng, Weili and Xing, Yitong and Liu, Fulong and Yang, Chengqun and Xiang, Antao and Tian, Feng and Gao, Jingnan and Cai, Jisong and Wang, Xin and Wu, Xiaomin and others},
  journal={arXiv preprint arXiv:2607.26657},
  year={2026}
}

@article{yan2026robust,
  title={Robust-WAM: Bridging Generative Pretraining and Semantic Foresight in World-Action Models},
  author={Yan, Haodong and Li, Junfeng and He, Junjie and Zhong, Zhide and Yu, MingMing and Song, Wenxuan and Zhu, Jiaguan and Zheng, Yangyang and Du, Yuqiao and You, Jiadi and others},
  journal={arXiv preprint arXiv:2608.05903},
  year={2026}
}

@inproceedings{bi2026motus,
  title={Motus: A unified latent action world model},
  author={Bi, Hongzhe and Tan, Hengkai and Xie, Shenghao and Wang, Zeyuan and Huang, Shuhe and Liu, Haitian and Zhao, Ruowen and Feng, Yao and Xiang, Chendong and Rong, Yinze and others},
  booktitle={Proceedings of the IEEE/CVF Conference on Computer Vision and Pattern Recognition},
  pages={35101--35113},
  year={2026}
}

@article{bjorck2025gr00t,
  title={Gr00t n1: An open foundation model for generalist humanoid robots},
  author={Bjorck, Johan and Casta{\~n}eda, Fernando and Cherniadev, Nikita and Da, Xingye and Ding, Runyu and Fan, Linxi and Fang, Yu and Fox, Dieter and Hu, Fengyuan and Huang, Spencer and others},
  journal={arXiv preprint arXiv:2503.14734},
  year={2025}
}

@article{black2024pi_0,
  title = {{$\pi_0$}: A Vision-Language-Action Flow Model for General Robot Control},
  author = {Black, Kevin and Brown, Noah and Driess, Danny and Esmail, Adnan and Equi, Michael and Finn, Chelsea and Fusai, Niccolo and Groom, Lachy and Hausman, Karol and Ichter, Brian and others},
  journal = {arXiv preprint arXiv:2410.24164},
  year = {2024}
}

@article{bu2025univla,
  title={Univla: Learning to act anywhere with task-centric latent actions},
  author={Bu, Qingwen and Yang, Yanting and Cai, Jisong and Gao, Shenyuan and Ren, Guanghui and Yao, Maoqing and Luo, Ping and Li, Hongyang},
  journal={arXiv preprint arXiv:2505.06111},
  year={2025}
}

@article{cen2025worldvla,
  title={Worldvla: Towards autoregressive action world model},
  author={Cen, Jun and Yu, Chaohui and Yuan, Hangjie and Jiang, Yuming and Huang, Siteng and Guo, Jiayan and Li, Xin and Song, Yibing and Luo, Hao and Wang, Fan and others},
  journal={arXiv preprint arXiv:2506.21539},
  year={2025}
}

@article{chen2026lawam,
  title={Lawam: Latent world action models for efficient dynamics-aware robot policies},
  author={Chen, Jialei and Wang, Kai and Chen, Kang and Chen, Shuaihang and Gao, Feng and Tang, Wenhao and Li, Zhiyuan and Liu, Weilin and Yao, Zhuyu and Li, Boxun and others},
  journal={arXiv preprint arXiv:2606.15768},
  year={2026}
}

@article{chi2025diffusion,
  title={Diffusion policy: Visuomotor policy learning via action diffusion},
  author={Chi, Cheng and Xu, Zhenjia and Feng, Siyuan and Cousineau, Eric and Du, Yilun and Burchfiel, Benjamin and Tedrake, Russ and Song, Shuran},
  journal={The International Journal of Robotics Research},
  volume={44},
  number={10-11},
  pages={1684--1704},
  year={2025},
  publisher={Sage Publications Sage UK: London, England}
}

@article{fei2025libero,
  title={Libero-plus: In-depth robustness analysis of vision-language-action models},
  author={Fei, Senyu and Wang, Siyin and Shi, Junhao and Dai, Zihao and Cai, Jikun and Qian, Pengfang and Ji, Li and He, Xinzhe and Zhang, Shiduo and Fei, Zhaoye and others},
  journal={arXiv preprint arXiv:2510.13626},
  year={2025}
}

@article{kim2024openvla,
  title={Openvla: An open-source vision-language-action model},
  author={Kim, Moo Jin and Pertsch, Karl and Karamcheti, Siddharth and Xiao, Ted and Balakrishna, Ashwin and Nair, Suraj and Rafailov, Rafael and Foster, Ethan and Lam, Grace and Sanketi, Pannag and others},
  journal={arXiv preprint arXiv:2406.09246},
  year={2024}
}

@article{li2503unified,
  title={Unified video action model, 2025},
  author={Li, Shuang and Gao, Yihuai and Sadigh, Dorsa and Song, Shuran},
  journal={arXiv preprint arXiv:2503.00200},
  year={2025}
}

@article{lipman2022flow,
  title={Flow matching for generative modeling},
  author={Lipman, Yaron and Chen, Ricky TQ and Ben-Hamu, Heli and Nickel, Maximilian and Le, Matt},
  journal={arXiv preprint arXiv:2210.02747},
  year={2022}
}

@article{liu2023libero,
  title={Libero: Benchmarking knowledge transfer for lifelong robot learning},
  author={Liu, Bo and Zhu, Yifeng and Gao, Chongkai and Feng, Yihao and Liu, Qiang and Zhu, Yuke and Stone, Peter},
  journal={Advances in Neural Information Processing Systems},
  volume={36},
  pages={44776--44791},
  year={2023}
}

@article{pertsch2025fast,
  title={Fast: Efficient action tokenization for vision-language-action models},
  author={Pertsch, Karl and Stachowicz, Kyle and Ichter, Brian and Driess, Danny and Nair, Suraj and Vuong, Quan and Mees, Oier and Finn, Chelsea and Levine, Sergey},
  journal={arXiv preprint arXiv:2501.09747},
  year={2025}
}

@article{intelligence2025pi_,
  title = {{$\pi_{0.5}$}: A Vision-Language-Action Model with Open-World Generalization},
  author = {Intelligence, Physical and Black, Kevin and Brown, Noah and Darpinian, James and Dhabalia, Karan and Driess, Danny and Esmail, Adnan and Equi, Michael and Finn, Chelsea and Fusai, Niccolo and others},
  journal = {arXiv preprint arXiv:2504.16054},
  year = {2025}
}

@article{ye2026world,
  title={World action models are zero-shot policies},
  author={Ye, Seonghyeon and Ge, Yunhao and Zheng, Kaiyuan and Gao, Shenyuan and Yu, Sihyun and Kurian, George and Indupuru, Suneel and Tan, You Liang and Zhu, Chuning and Xiang, Jiannan and others},
  journal={arXiv preprint arXiv:2602.15922},
  year={2026}
}

@article{yuan2026fast,
  title={Fast-wam: Do world action models need test-time future imagination?},
  author={Yuan, Tianyuan and Dong, Zibin and Liu, Yicheng and Zhao, Hang},
  journal={arXiv preprint arXiv:2603.16666},
  year={2026}
}

@article{zhu2025unified,
  title={Unified world models: Coupling video and action diffusion for pretraining on large robotic datasets},
  author={Zhu, Chuning and Yu, Raymond and Feng, Siyuan and Burchfiel, Benjamin and Shah, Paarth and Gupta, Abhishek},
  journal={arXiv preprint arXiv:2504.02792},
  year={2025}
}

@article{brohan2023rt,
  title={Rt-2: Vision-language-action models transfer web knowledge to robotic control},
  author={Brohan, Anthony and Brown, Noah and Carbajal, Justice and Chebotar, Yevgen and Chen, Xi and Choromanski, Krzysztof and Ding, Tianli and Driess, Danny and Dubey, Avinava and Finn, Chelsea and others},
  journal={arXiv preprint arXiv:2307.15818},
  year={2023}
}

@inproceedings{du2024video,
  title={Video language planning},
  author={Du, Yilun and Yang, Sherry and Florence, Pete and Xia, Fei and Wahid, Ayzaan and Sermanet, Pierre and Yu, Tianhe and Abbeel, Pieter and Tenenbaum, Joshua B and Kaelbling, Leslie and others},
  booktitle={International Conference on Learning Representations},
  volume={2024},
  pages={31138--31155},
  year={2024}
}

@article{guo2024prediction,
  title={Prediction with action: Visual policy learning via joint denoising process},
  author={Guo, Yanjiang and Hu, Yucheng and Zhang, Jianke and Wang, Yen-Jen and Chen, Xiaoyu and Lu, Chaochao and Chen, Jianyu},
  journal={Advances in Neural Information Processing Systems},
  volume={37},
  pages={112386--112410},
  year={2024}
}

@inproceedings{bharadhwaj2024track2act,
  title={Track2act: Predicting point tracks from internet videos enables generalizable robot manipulation},
  author={Bharadhwaj, Homanga and Mottaghi, Roozbeh and Gupta, Abhinav and Tulsiani, Shubham},
  booktitle={European Conference on Computer Vision},
  pages={306--324},
  year={2024},
  organization={Springer}
}

@inproceedings{ko2024learning,
  title={Learning to act from actionless videos through dense correspondences},
  author={Ko, Po-Chen and Mao, Jiayuan and Du, Yilun and Sun, Shao-Hua and Tenenbaum, Joshua B},
  booktitle={International Conference on Learning Representations},
  volume={2024},
  pages={40938--40958},
  year={2024}
}

@article{shen2026videovla,
  title={Videovla: Video generators can be generalizable robot manipulators},
  author={Shen, Yichao and Wei, Fangyun and Du, Zhiying and Liang, Yaobo and Lu, Yan and Yang, Jiaolong and Zheng, Nanning and Guo, Baining},
  journal={Advances in neural information processing systems},
  volume={38},
  pages={95597--95621},
  year={2026}
}

@article{xu2024flow,
  title={Flow as the cross-domain manipulation interface},
  author={Xu, Mengda and Xu, Zhenjia and Xu, Yinghao and Chi, Cheng and Wetzstein, Gordon and Veloso, Manuela and Song, Shuran},
  journal={arXiv preprint arXiv:2407.15208},
  year={2024}
}

@article{zhi20253dflowaction,
  title={3dflowaction: Learning cross-embodiment manipulation from 3d flow world model},
  author={Zhi, Hongyan and Chen, Peihao and Zhou, Siyuan and Dong, Yubo and Wu, Quanxi and Han, Lei and Tan, Mingkui},
  journal={arXiv preprint arXiv:2506.06199},
  year={2025}
}

@article{lou2026mask,
  title={Mask World Model: Predicting What Matters for Robust Robot Policy Learning},
  author={Lou, Yunfan and Chi, Xiaowei and Zhang, Xiaojie and Qian, Zezhong and Li, Chengxuan and Zhang, Rongyu and Lyu, Yaoxu and Song, Guoyu and Fu, Chuyao and Xu, Haoxuan and others},
  journal={arXiv preprint arXiv:2604.19683},
  year={2026}
}

@inproceedings{huang2024ardup,
  title={Ardup: Active region video diffusion for universal policies},
  author={Huang, Shuaiyi and Levy, Mara and Jiang, Zhenyu and Anandkumar, Anima and Zhu, Yuke and Fan, Linxi and Huang, De-An and Shrivastava, Abhinav},
  booktitle={2024 IEEE/RSJ International Conference on Intelligent Robots and Systems (IROS)},
  pages={8465--8472},
  year={2024},
  organization={IEEE}
}

@article{yan2026s,
  title={S-vam: Shortcut video-action model by self-distilling geometric and semantic foresight},
  author={Yan, Haodong and Zhong, Zhide and Zhu, Jiaguan and He, Junjie and Yuan, Weilin and Song, Wenxuan and Gong, Xin and Cai, Yingjie and Zhao, Guanyi and Yan, Xu and others},
  journal={arXiv preprint arXiv:2603.16195},
  year={2026}
}

@article{cheang2024gr,
  title={Gr-2: A generative video-language-action model with web-scale knowledge for robot manipulation},
  author={Cheang, Chi-Lam and Chen, Guangzeng and Jing, Ya and Kong, Tao and Li, Hang and Li, Yifeng and Liu, Yuxiao and Wu, Hongtao and Xu, Jiafeng and Yang, Yichu and others},
  journal={arXiv preprint arXiv:2410.06158},
  year={2024}
}

@article{cen2025rynnvla,
  title={Rynnvla-002: A unified vision-language-action and world model},
  author={Cen, Jun and Huang, Siteng and Yuan, Yuqian and Li, Kehan and Yuan, Hangjie and Yu, Chaohui and Hou, Bohan and Jiang, Yuming and Guo, Jiayan and Li, Xin and others},
  journal={arXiv preprint arXiv:2511.17502},
  year={2025}
}

@article{kim2026cosmos,
  title={Cosmos policy: Fine-tuning video models for visuomotor control and planning},
  author={Kim, Moo Jin and Gao, Yihuai and Lin, Tsung-Yi and Lin, Yen-Chen and Ge, Yunhao and Lam, Grace and Liang, Percy and Song, Shuran and Liu, Ming-Yu and Finn, Chelsea and others},
  journal={arXiv preprint arXiv:2601.16163},
  year={2026}
}

@article{han2024dual,
  title={A dual process vla: Efficient robotic manipulation leveraging vlm},
  author={Han, ByungOk and Kim, Jaehong and Jang, Jinhyeok},
  journal={arXiv preprint arXiv:2410.15549},
  year={2024}
}

@article{won2025dual,
  title={Dual-stream diffusion for world-model augmented vision-language-action model},
  author={Won, John and Lee, Kyungmin and Jang, Huiwon and Kim, Dongyoung and Shin, Jinwoo},
  journal={arXiv preprint arXiv:2510.27607},
  year={2025}
}

@article{yang2025covar,
  title={CoVAR: Co-generation of Video and Action for Robotic Manipulation via Multi-Modal Diffusion},
  author={Yang, Liudi and Bai, Yang and Eskandar, George and Shen, Fengyi and Altillawi, Mohammad and Chen, Dong and Liu, Ziyuan and Valada, Abhinav},
  journal={arXiv preprint arXiv:2512.16023},
  year={2025}
}

@inproceedings{chen2026unified,
  title={Unified diffusion VLA: Vision-language-action model via joint discrete denosing diffusion process},
  author={Chen, Jiayi and Song, Wenxuan and Ding, Pengxiang and Zhou, Ziyang and Zhao, Han and Tang, Barrett and Wang, Donglin and Li, Haoang},
  booktitle={International Conference on Learning Representations},
  volume={2026},
  pages={139291--139311},
  year={2026}
}

@article{li2026world,
  title={World-value-action model: Implicit planning for vision-language-action systems},
  author={Li, Runze and Zhang, Hongyin and Jin, Junxi and Zeng, Qixin and Zhuang, Zifeng and Tang, Yiqi and Lyu, Shangke and Wang, Donglin},
  journal={arXiv preprint arXiv:2604.14732},
  year={2026}
}

@article{zhao2026frappe,
  title={Frappe: Infusing world modeling into generalist policies via multiple future representation alignment},
  author={Zhao, Han and Wang, Jingbo and Song, Wenxuan and Chen, Shuai and Liu, Yang and Wang, Yan and Li, Haoang and Wang, Donglin},
  journal={arXiv preprint arXiv:2602.17259},
  year={2026}
}

@article{yuan2026adaworldpolicy,
  title={AdaWorldPolicy: World-Model-Driven Diffusion Policy with Online Adaptive Learning for Robotic Manipulation},
  author={Yuan, Ge and Qiao, Qiyuan and Zhang, Jing and Xu, Dong},
  journal={arXiv preprint arXiv:2602.20057},
  year={2026}
}

@inproceedings{zheng2026x,
  title={X-vla: Soft-prompted transformer as scalable cross-embodiment vision-language-action model},
  author={Zheng, Jinliang and Li, Jianxiong and Wang, Zhihao and Liu, Dongxiu and Kang, Xirui and Feng, Yuchun and Zheng, Yinan and Zou, Jiayin and Chen, Yilun and Zeng, Jia and others},
  booktitle={International Conference on Learning Representations},
  volume={2026},
  pages={60580--60606},
  year={2026}
}

@article{yang2026abot,
  title={Abot-m0: Vla foundation model for robotic manipulation with action manifold learning},
  author={Yang, Yandan and Zeng, Shuang and Lin, Tong and Chang, Xinyuan and Qi, Dekang and Xiao, Junjin and Liu, Haoyun and Chen, Ronghan and Chen, Yuzhi and Huo, Dongjie and others},
  journal={arXiv preprint arXiv:2602.11236},
  year={2026}
}

@article{zheng2025flare,
  title={Flare: Robot learning with implicit world modeling},
  author={Zheng, Ruijie and Wang, Jing and Reed, Scott and Bjorck, Johan and Fang, Yu and Hu, Fengyuan and Jang, Joel and Kundalia, Kaushil and Lin, Zongyu and Magne, Loic and others},
  journal={arXiv preprint arXiv:2505.15659},
  year={2025}
}

@article{lyu2026lda,
  title={Lda-1b: Scaling latent dynamics action model via universal embodied data ingestion},
  author={Lyu, Jiangran and Liu, Kai and Zhang, Xuheng and Liao, Haoran and Feng, Yusen and Zhu, Wenxuan and Shen, Tingrui and Chen, Jiayi and Zhang, Jiazhao and Dong, Yifei and others},
  journal={arXiv preprint arXiv:2602.12215},
  year={2026}
}

@article{ye2026gigaworld,
  title={GigaWorld-Policy: An Efficient Action-Centered World--Action Model},
  author={Ye, Angen and Wang, Boyuan and Ni, Chaojun and Huang, Guan and Zhao, Guosheng and Li, Hao and Li, Hengtao and Li, Jie and Lv, Jindi and Liu, Jingyu and others},
  journal={arXiv preprint arXiv:2603.17240},
  year={2026}
}

@inproceedings{zhao2025cot,
  title={Cot-vla: Visual chain-of-thought reasoning for vision-language-action models},
  author={Zhao, Qingqing and Lu, Yao and Kim, Moo Jin and Fu, Zipeng and Zhang, Zhuoyang and Wu, Yecheng and Li, Zhaoshuo and Ma, Qianli and Han, Song and Finn, Chelsea and others},
  booktitle={2025 IEEE/CVF Conference on Computer Vision and Pattern Recognition (CVPR)},
  pages={1702--1713},
  year={2025},
  organization={IEEE}
}

@inproceedings{wu2024unleashing,
  title={Unleashing large-scale video generative pre-training for visual robot manipulation},
  author={Wu, Hongtao and Jing, Ya and Cheang, Chilam and Chen, Guangzeng and Xu, Jiafeng and Li, Xinghang and Liu, Minghuan and Li, Hang and Kong, Tao},
  booktitle={International Conference on Learning Representations},
  volume={2024},
  pages={10641--10662},
  year={2024}
}

@article{team2024octo,
  title={Octo: An open-source generalist robot policy},
  author={Team, Octo Model and Ghosh, Dibya and Walke, Homer and Pertsch, Karl and Black, Kevin and Mees, Oier and Dasari, Sudeep and Hejna, Joey and Kreiman, Tobias and Xu, Charles and others},
  journal={arXiv preprint arXiv:2405.12213},
  year={2024}
}

@inproceedings{liu2025rdt,
  title={Rdt-1b: a diffusion foundation model for bimanual manipulation},
  author={Liu, Songming and Wu, Lingxuan and Li, Bangguo and Tan, Hengkai and Chen, Huayu and Wang, Zhengyi and Xu, Ke and Su, Hang and Zhu, Jun},
  booktitle={International Conference on Learning Representations},
  volume={2025},
  pages={29982--30009},
  year={2025}
}

@article{brohan2022rt,
  title={Rt-1: Robotics transformer for real-world control at scale},
  author={Brohan, Anthony and Brown, Noah and Carbajal, Justice and Chebotar, Yevgen and Dabis, Joseph and Finn, Chelsea and Gopalakrishnan, Keerthana and Hausman, Karol and Herzog, Alex and Hsu, Jasmine and others},
  journal={arXiv preprint arXiv:2212.06817},
  year={2022}
}

@article{team2026kairos,
  title={Kairos: A Native World Model Stack for Physical AI},
  author={Team, Kairos and Wang, Fei and You, Shan and Zhang, Qiming and Huang, Tao and Fu, Zuoyi and Zheng, Zhisheng and Xi, Yunlong and Lv, Feng and Wu, Xiaoming and others},
  journal={arXiv preprint arXiv:2606.16533},
  year={2026}
}

@article{ma2026dit4dit,
  title={Dit4dit: Jointly modeling video dynamics and actions for generalizable robot control},
  author={Ma, Teli and Zheng, Jia and Wang, Zifan and Jiang, Chunli and Cui, Andy and Liang, Junwei and Yang, Shuo},
  journal={arXiv preprint arXiv:2603.10448},
  year={2026}
}

@article{jang2025dreamgen,
  title={Dreamgen: Unlocking generalization in robot learning through video world models},
  author={Jang, Joel and Ye, Seonghyeon and Lin, Zongyu and Xiang, Jiannan and Bjorck, Johan and Fang, Yu and Hu, Fengyuan and Huang, Spencer and Kundalia, Kaushil and Lin, Yen-Chen and others},
  journal={arXiv preprint arXiv:2505.12705},
  year={2025}
}

@article{li2026cogvla,
  title={CogVLA: Cognition-aligned vision-language-action models via instruction-driven routing \& sparsification},
  author={Li, Wei and Zhang, Renshan and Shao, Rui and He, Jie and Nie, Liqiang},
  journal={Advances in neural information processing systems},
  volume={38},
  pages={137646--137675},
  year={2026}
}

@misc{kairosteam2026kairosregretawarenativeworldaction,
      title={Kairos: A Regret-Aware Native World-Action Model Stack for Physical AI}, 
      author={Kairos Team and Fei Wang and Shan You and Qiming Zhang and Tao Huang and Zuoyi Fu and Zhisheng Zheng and Yunlong Xi and Feng Lv and Xiaoming Wu and Zeyu Liu and Cong Wan and Pu Li and Ruiqing Yang and Xiaoou Li and Wei Wang and Kangkang Zhu and Yuwei Zhang and Shi Fu and Zheng Zhang and Xiaoning Wu and Xuzeng Fan and Dacheng Tao and Xiaogang Wang},
      year={2026},
      eprint={2606.16533},
      archivePrefix={arXiv},
      primaryClass={cs.AI},
      url={https://arxiv.org/abs/2606.16533}, 
}

@article{world_model_benchmarks_survey_2026,
  title   = {A Survey of World Model Benchmarks},
  author  = {Zheng Zhang and Zhixiang Guo and Yuan Si and Siyuan Liang and Shunyu Liu and Weibo Gao and Song Wang and Leszek Rutkowski and Giuseppe Valenzise and Ming-Hsuan Yang and Lin William Cong and M. Jamal Deen and Sally Cripps and Dacheng Tao},
  year    = {2026},
  note    = {Manuscript}
}
\endgroup

\clearpage
\appendix
\section{Training Details}
\label{app:training_details}
\paragraph{Architecture and inputs.}
We initialize the visual branch from Wan2.1-T2V-1.3B, retaining its video DiT,
text encoder, and video VAE~\citep{wan2025wan}. We precompute instruction
embeddings with the corresponding Wan2.1 text encoder. The video DiT comprises
30 transformer blocks, while the default action expert uses one block and
attends to the final-layer video K/V cache. We additionally evaluate action
experts with 6, 12, and 30 blocks in the depth ablation; the component and
register--teacher ablations use 30-block action experts. The video branch uses hidden
dimension $d_v=1536$, whereas the action expert uses $d_a=1024$ and is
initialized from a linearly interpolated Wan2.1 ActionDiT checkpoint. The
action horizon is $H=32$.

Each training example spans 33 observation time steps and a 32-step action
chunk. The data loader samples video frames at indices $0,4,\ldots,32$,
yielding nine RGB frames. The VAE then compresses these to
$1+(9-1)/4=3$ latent temporal positions: one current-frame position and two
future slots. The teacher observation pair uses the first and last sampled
frames, corresponding to original indices 0 and 32. For LIBERO, we concatenate the two synchronized camera views along the image width
before VAE encoding, producing a $224\times448$ image composed of two
$224\times224$ views. The policy additionally receives an 8-dimensional
proprioceptive state. Each action is seven-dimensional, comprising a 6-DoF
end-effector pose and one gripper-control dimension.

\paragraph{Optimization and inference.}
We jointly optimize the video expert, action expert, dynamics registers,
latent-action projection, and proprioceptive encoder using AdamW, with
the settings in Table~\ref{tab:shared_hyperparameters}. Video and action
training use continuous flow matching. Latent-action supervision uses
mean-squared error against detached targets from the frozen LaWM teacher.
We disable readability registers and allow action gradients to propagate
through the video K/V interface. At inference, one video prefill at
$\sigma=1.0$ supplies the cache reused throughout action denoising.
None of the reported variants receives additional policy-level embodied pretraining before LIBERO training.

\begin{table}[ht]
    \centering
    \footnotesize
    \caption{Shared training and inference hyperparameters.}
    \label{tab:shared_hyperparameters}
    \begin{tabular}{ll}
        \toprule
        Parameter & Value \\
        \midrule
        Optimizer & AdamW \\
        Learning rate & $1\times10^{-4}$ \\
        Weight decay & $0.01$ \\
        Learning-rate schedule & Cosine decay \\
        Maximum gradient norm & $1.0$ \\
        Flow schedules & 1,000 timesteps; shift 5.0 \\
        Video / action / latent-action loss weights & $1.0$ / $1.0$ / $1.0$ \\
        Dynamics registers / teacher target dimension & 16 / 32 \\
        Action horizon / inference denoising steps & 32 / 10 \\
        Training seed & 42 \\
        \bottomrule
    \end{tabular}
\end{table}

\paragraph{Training budgets.}
Table~\ref{tab:training_configurations} summarizes the dataset-specific
training configurations. Piper uses BF16 with an online FP32 LaWM teacher
and produces a checkpoint at 10,500 optimizer updates.

\begin{table}[ht]
    \centering
    \footnotesize
    \caption{Dataset-specific training configurations.}
    \label{tab:training_configurations}
    \begin{tabular}{lrrr}
        \toprule
        Setting & LIBERO & RoboCasa & Piper \\
        \midrule
        Epochs & 20 & 20 & 20 \\
        Global batch size & 128 & 128 & 128 \\
        \bottomrule
    \end{tabular}
\end{table}


\paragraph{Action-to-video layer mapping.}
Each action block reads K/V from a specified video block. With zero-based
video block indices, the mappings for action depths 1, 6, and 12 are
$[29]$, $[0,6,12,17,23,29]$, and
$[0,3,5,8,11,13,16,18,21,24,26,29]$, respectively. The 30-block action
expert uses the identity mapping $[0,1,\ldots,29]$. Each depth uses its
corresponding interpolated Action DiT initialization. The video expert
retains all 30 blocks in every case.

\paragraph{Policy initialization and pretraining scope.}
The Video DiT is initialized from pretrained Wan2.1 weights, and the
Action DiT backbone is initialized from the same checkpoint through layer
mapping and weight interpolation. We do not perform an additional
large-scale robot-data pretraining stage for either branch before
task-specific training.

\paragraph{Separate training forward passes.}
For Future-KV with Video DiT registers, training requires two video forward
passes to prevent demonstrated future observations from leaking into the
action-conditioning features, matching the information available at inference.
The first combines the clean current latent with noised demonstrated
future latents and supplies the video loss and register states for LaWM
supervision. The second combines the clean current latent with newly sampled
Gaussian future slots at $\sigma=1.0$ and recomputes the register states and
layer-wise K/V, including register-token K/V, for the action loss. The default
one-layer Action DiT reads the final-layer cache. These passes share model
parameters and learnable register initializations, but do not reuse each
other's hidden states or caches. Demonstrated futures influence the action
policy through parameter learning, not as inputs to its conditioning pass.
Only the second video pass is required at inference.

\paragraph{Frozen LaWM teacher.}
We use the publicly released LaWM checkpoint from
LaWAM~\citep{chen2026lawam} as a frozen teacher for latent-action supervision.
The teacher provides training targets and is discarded at inference.
Its pretraining on external embodied data provides indirect access to
external embodied knowledge during policy training; this data provenance
remains relevant to comparisons even though the teacher is absent at
inference.

\paragraph{P.T. reporting convention.}
In the tables, P.T. denotes large-scale embodied-data pretraining of the
policy backbone or action expert before task-specific training, including
inherited weights from an embodied-pretrained policy. It excludes
general-purpose video pretraining and pretraining of auxiliary frozen
teachers. Under this convention, ForeWAM is marked as not using
policy-level embodied pretraining, while its use of the pretrained LaWM
teacher is disclosed above. This marking does not imply that the complete
training pipeline is free of external embodied knowledge.

\paragraph{Ablation baseline.}
The baseline with both components disabled in Table~\ref{tab:component_design}
is Fast-WAM retrained with Wan2.1 under the same experimental configuration as
the 30-layer ForeWAM ablations. It achieves 53.6\% overall success on
LIBERO-Plus and is distinct from the 6B Fast-WAM configuration reported in
Table~\ref{tab:libero_plus_results} (51.5\%).

\section{Details of the Latency Comparison}
\label{app:latency_comparison}

\paragraph{Evaluation setting.}
The comparison in Table~\ref{tab:inference_latency} measures the time required
for one policy query on LIBERO-10 using a single NVIDIA A800 GPU with
80\,GB of memory. Both methods use 10 action-denoising steps. We perform
5 warm-up iterations before collecting 10 timed runs and report the
arithmetic mean in milliseconds. We time action-chunk prediction,
excluding environment stepping, rendering, and episode resets.

\paragraph{ForeWAM configuration.}
ForeWAM uses the Wan2.1-T2V-1.3B visual branch and the one-layer Action DiT
described in Appendix~\ref{app:training_details}. The two camera views form
a $224\times448$ input, accompanied by the current proprioceptive state.
The model encodes the current observation and performs one Video DiT
prefill over the current latent, noisy future slots, and dynamics registers.
The action expert reuses the final-layer video K/V cache across all 10
denoising steps to produce a 32-action chunk. Readability registers are
disabled. Neither future observations nor the latent-action teacher is
required at inference, and no future video is decoded.

\paragraph{Timing breakdown.}
\textit{VAE Enc.} measures encoding of the current visual observation into
latent space. \textit{Visual} covers visual-context computation, including
ForeWAM's Video DiT prefill and construction of the K/V cache.
\textit{Action} covers action generation across the full denoising loop.
\textit{Other} accounts for
the remaining inference overhead outside these three stages, while
\textit{Overall} reports the total policy-query latency. For the reported
means, these components sum to 88.7\,ms for ForeWAM and 556.3\,ms for
Fast-WAM.

\paragraph{Latency reduction.}
ForeWAM reduces total latency by 84.1\%, corresponding to a
$6.27\times$ speedup. Action generation accounts for 448.9\,ms of the
467.6\,ms total saving, whereas visual-context computation differs by only
1.3\,ms.


\section{Real-World Experiment Details}
\label{app:real_world_details}

\paragraph{Platform and task definitions.}
We use a 6-DoF AgileX Piper arm with an external RGB camera and a
wrist-mounted RGB camera. The four tasks follow the naming and order in
the main text. \textbf{Place Pen in Cup (T1)} requires placing the pen in
the target cup. \textbf{Place Carrot in Bowl (T2)} requires picking up the
carrot and placing it in the bowl. \textbf{Place Blocks in Drawer and
Close It (T3)} requires placing the blocks in the second drawer and then
closing it. \textbf{Stack Three Bowls (T4)} requires forming a complete
stack of three bowls.

\paragraph{Dataset and observation format.}
The dataset contains 125 demonstrations and 71,702 recorded frames
collected through master--slave teleoperation. We train a single
language-conditioned policy jointly on demonstrations from all four tasks.

The external and wrist cameras provide RGB images at
$480\times640$ resolution. Each view is
resized to $224\times224$ and the two views are concatenated horizontally
to form a $224\times448$ video input. Both recorded proprioceptive states
and actions are seven-dimensional. The Piper configuration preserves the
recorded absolute action representation without converting it to a
state-relative delta. State and action normalization uses the dataset's
1st and 99th percentile statistics. A training window contains 32 actions;
its video sampling and temporal compression follow
Appendix~\ref{app:training_details}. Instruction embeddings are cached
with a maximum context length of 128 tokens.

The policy uses the default one-block action expert and 16 dynamics
registers described in Appendix~\ref{app:training_details}. Online LaWM
supervision uses only the external-camera image pair; the teacher does
not receive robot states or actions.

\paragraph{Evaluation protocol.}
The ID evaluation protocol specifies 10 trials per task, or 40 trials per
method. Object placements are randomized across trials and matched
across methods, and success rates are reported separately for each
task according to the task goals above. At inference, the model uses the current
two-camera observation and proprioceptive state, performs one video
prefill with pure-noise future slots, and generates a 32-action chunk
using 10 action-denoising steps. Demonstrated future observations and
the LaWM teacher are not used during policy execution.

\section{Detailed Component Ablation}
\label{app:component_details}

Table~\ref{tab:component_details} provides the complete seven-category
results for the component ablation in Section~\ref{sec:motivation_context}.
All configurations use a 30-layer Wan2.1 Video DiT and a 30-layer Action
DiT without additional policy-level embodied pretraining. The results match
 Table~\ref{tab:component_design}.

\begin{table}[H]
    \centering
    \footnotesize
    \setlength{\tabcolsep}{3pt}
    \caption{Complementarity of Future-KV and latent-action supervision.
    LIBERO-Plus success rate (\%), without embodied pretraining, using a
    30-layer Action DiT. The baseline is Fast-WAM retrained with Wan2.1
    under matched settings. $\checkmark$/$\times$ indicate inclusion/omission.
    Bold marks the best result in each column.}
    \label{tab:component_details}
    \begin{tabular}{@{}ll|cccccccc@{}}
        \toprule
        Latent action & Future feature & Camera & Robot & Language & Light
        & Background & Noise & Layout & Avg \\
        \midrule
        $\times$ & $\times$
        & 37.0 & 36.5 & 64.1 & 81.5 & 49.3 & 50.1 & 63.7 & 53.6 \\
        $\checkmark$ & $\times$
        & 49.0 & 36.8 & 65.1 & 75.9 & 48.5 & 50.8 & 61.7 & 54.8 \\
        $\times$ & $\checkmark$
        & 37.8 & 41.0 & 63.9 & 83.4 & 65.7 & 61.5 & 65.6 & 58.5 \\
        $\checkmark$ & $\checkmark$
        & \textbf{64.6} & \textbf{61.6} & \textbf{86.6} & \textbf{95.3}
        & \textbf{74.1} & \textbf{77.7} & \textbf{81.7} & \textbf{76.7} \\
        \bottomrule
    \end{tabular}
\end{table}

\noindent\begin{minipage}{\linewidth}
\section{Detailed LIBERO Results}
\label{app:libero_results}

Table~\ref{tab:libero_results_detailed} reports the suite-level results
corresponding to the average success rates in Table~\ref{tab:libero_results}.

\begin{table}[H]
    \centering
    \caption{Detailed success rates (\%) on the four LIBERO suites.}
    \label{tab:libero_results_detailed}
    \footnotesize
    \setlength{\tabcolsep}{1.5pt}
    \renewcommand{\arraystretch}{1.08}
    \begin{tabularx}{\linewidth}{@{}l|c|*{4}{>{\centering\arraybackslash}X}|>{\centering\arraybackslash}X@{}}
        \toprule
        Method & P.T. & Spatial & Object & Goal & Long & \textbf{Avg.} \\
        \midrule
        OpenVLA~\citep{kim2024openvla} & $\checkmark$ & 84.7 & 88.4 & 79.2 & 53.7 & 76.5 \\
        WorldVLA~\citep{cen2025worldvla} & $\checkmark$ & 87.6 & 96.2 & 83.4 & 60.0 & 81.8 \\
        $\pi_0$-Fast~\citep{pertsch2025fast} & $\checkmark$ & 96.4 & 96.8 & 88.6 & 60.2 & 85.5 \\
        $\pi_0$~\citep{black2024pi_0} & $\checkmark$ & 96.8 & 98.8 & 95.8 & 85.2 & 94.1 \\
        $\pi_{0.5}$~\citep{intelligence2025pi_} & $\checkmark$ & \textbf{98.8} & 98.2 & \textbf{98.0} & 92.4 & 96.9 \\
        UniVLA~\citep{bu2025univla} & $\checkmark$ & 96.5 & 96.8 & 95.6 & 92.0 & 95.2 \\
        CogVLA~\citep{li2026cogvla}  & $\checkmark$ & \underline{98.6} & 98.8 & 96.6 & \underline{95.4} & 97.4 \\
        Fast-WAM~\citep{yuan2026fast} & $\times$ & 98.2 & \textbf{100.0} & \underline{97.0} & 95.2 & \underline{97.6} \\
        \rowcolor{oursrow}
        \textbf{ForeWAM} & $\times$ & 97.6 & \underline{99.8} & \textbf{98.0} & \textbf{96.2} & \textbf{97.9} \\
        \bottomrule
    \end{tabularx}
\end{table}
\end{minipage}

\section{Benchmark Evaluation Protocols}
\label{app:evaluation_protocols}

\paragraph{LIBERO and LIBERO-Plus.}
We evaluate our model on LIBERO~\citep{liu2023libero} and LIBERO-Plus~\citep{fei2025libero}. For LIBERO,
we follow the standard benchmarking protocol and train on the four standard suites: Spatial, Object, Goal,
and LIBERO-Long. Each suite contains 500 expert demonstrations across 10 tasks.

LIBERO-Plus extends the LIBERO tasks with additional visual and layout variations. For training, we use the
original LIBERO demonstrations and exclude the augmented LIBERO-Plus training data. We evaluate the trained
policies under the LIBERO-Plus protocol and report overall success rate.
For every table reporting Avg, we aggregate evaluation episodes as
\begin{equation}
    \mathrm{Avg}=\mathrm{Overall\ SR}
    =100\times\frac{\sum_c S_c}{\sum_c N_c},
\end{equation}
where $S_c$ and $N_c$ are the numbers of successful and total evaluation
 episodes in group $c$, respectively (e.g., a LIBERO suite or a LIBERO-Plus
 perturbation category). Thus, Avg weights each group by its episode count;
it is not the unweighted arithmetic mean of category percentages unless
all groups contain the same number of episodes. Aggregation uses the
underlying counts before rounding.

\paragraph{RoboCasa.}
 We evaluate on 24 static kitchen manipulation tasks with a single Franka Emika Panda arm. Each task uses five scenes with different floor plans and visual styles, with 10 trials per scene. We report the mean success rate over all 24 tasks and three random seeds, for 3{,}600 trials in total. The evaluation uses unseen object instances, and two of the five scenes for each task contain styles absent from training.

We train ForeWAM on the successful human demonstrations in the released RoboCasa dataset and do not use the additional MimicGen demonstrations.

\paragraph{Real-world Experiments.}
Our real-world setup consists of a 6-DoF AgileX Piper arm with third-person and wrist-mounted RGB cameras. The task suite includes four tabletop tasks: \textbf{Place Pen in Cup (T1)}, \textbf{Place Carrot in Bowl (T2)}, \textbf{Place Blocks in Drawer and Close It (T3)}, and \textbf{Stack Three Bowls (T4)}. These tasks cover precision placement, pick-and-place, articulated-object manipulation, and multi-stage stacking. In T3, the target is the second drawer. The training dataset contains 125 robot demonstrations collected through master--slave teleoperation, totaling 71{,}702 frames across the four tasks. We train ForeWAM jointly on demonstrations from all four tasks. The ID evaluation protocol specifies 10 trials per task, with object placements randomized across trials and matched across methods. We also evaluate out-of-distribution (OOD) robustness by perturbing object attributes in pen placement and drawer manipulation (Figure~\ref{fig:real_robot}).

\end{document}